\documentclass[a4paper]{spie}  
\usepackage[]{graphicx}

\usepackage{multirow}
\usepackage{rotating}
\usepackage{color}

\usepackage[numbers,sort]{natbib}

\makeatletter
\def\Hline{%
\noalign{\ifnum0=`}\fi\hrule \@height 1pt \futurelet
\reserved@a\@xhline}
\makeatother

\title{All-in-one platform for AI R\&D in medical imaging, encompassing data collection, selection, annotation, and pre-processing} 

\author{Changhee Han$^a$, Kyohei Shibano$^b$, Wataru Ozaki$^a$, Keishiro Osaki$^a$, Takafumi Haraguchi$^c$, Daisuke Hirahara$^d$, Shumon Kimura$^e$, Yasuyuki Kobayashi$^f$ and Gento Mogi$^b$
\skiplinehalf
$^a$Callisto Inc., Toshima, Tokyo, Japan\\[0mm]
$^b$The University of Tokyo, Depart. Technology Management for Innovation, Bunkyo, Tokyo, Japan\\[0mm]
$^c$St. Marianna University School of Medicine, Depart. Advanced Biomedical Imaging and Informatics, Kawasaki, Kanagawa, Japan\\[0mm]
$^d$Harada Academy, Depart. AI Research Lab, Kagoshima, Kagoshima, Japan\\[0mm]
$^e$St. Marianna University, Yokohama Seibu Hospital Clinical Training Center, Yokohama, Kanagawa, Japan\\[0mm]
$^f$St. Marianna University School of Medicine, Depart. Medical Information and Communication Technology Research, Kawasaki, Kanagawa, Japan}

\begin{document} 
  \maketitle 

\begin{abstract}
Deep Learning is advancing medical imaging Research and Development (R\&D), leading to the frequent clinical use of Artificial Intelligence/Machine Learning (AI/ML)-based medical devices. However, to advance AI R\&D, two challenges arise: 1) significant data imbalance, with most data from Europe/America and under 10\% from Asia, despite its 60\% global population share; and 2) hefty time and investment needed to curate proprietary datasets for commercial use. In response, we established the first commercial medical imaging platform, encompassing steps like: 1) data collection, 2) data selection, 3) annotation, and 4) pre-processing. Moreover, we focus on harnessing under-represented data from Japan and broader Asia, including Computed Tomography, Magnetic Resonance Imaging, and Whole Slide Imaging scans. Using the collected data, we are preparing/providing ready-to-use datasets for medical AI R\&D by 1) offering these datasets to AI firms, biopharma, and medical device makers and 2) using them as training/test data to develop tailored AI solutions for such entities. We also aim to merge Blockchain for data security and plan to synthesize rare disease data $via$ generative AI.
\end{abstract}



\section{INTRODUCTION}
\label{sec:intro}

\begin{figure}
   \begin{center}
   \begin{tabular}{c}
   \includegraphics[width=16cm]{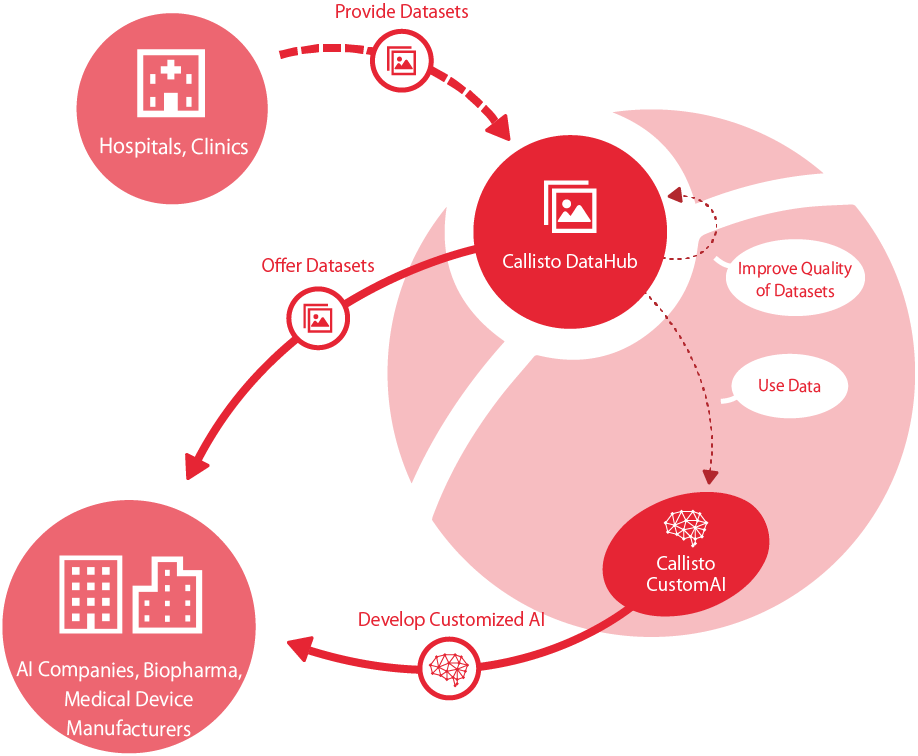}
   \end{tabular}
   \end{center}
   \caption[example] 
   { \label{fig:business_ecosystem} 
Overview of our business ecosystem: hospitals and clinics provide anonymized medical images and clinical data, possibly with their annotation, to a cloud platform; then, Callisto meticulously selects, annotates (with an expert radiologist/pathologist's rigorous double-check), and pre-processes them to prepare ready-to-use datasets for medical AI R\&D; finally, Callisto 1) offers these datasets to AI companies, biopharma, and medical device manufacturers, and 2) leverages them as training/test data to develop customized AI solutions for such entities.}
   \end{figure}

Deep Learning-based medical imaging Research and Development (R\&D), notably Computer Aided Diagnosis (CAD), is significantly increasing Artificial Intelligence/Machine Learning (AI/ML)-based medical devices in clinical use. Nevertheless, for more accurate and robust AI R\&D, two challenges remain: 1) significant data imbalance, with most data from Europe/America and under 10\% from Asia, despite its 60\% global population share; and 2) extensive time and resources required to prepare proprietary datasets for commercial use. Therefore, we launched the first commercial medical imaging platform, offering datasets after: 1) data collection, 2) data selection, 3) annotation, and 4) pre-processing. We particularly collect under-represented data from Japan and wider Asia, including Computed Tomography (CT), Magnetic Resonance Imaging (MRI), and Whole Slide Imaging (WSI) scans. Dr. Changhee Han, the founder of Callisto Inc., is among the leading young medical AI researchers in Japan. He has been pivotal in fostering AI-healthcare synergies in Japan using generative AI for Data Augmentation and physician training~\cite{han2020}, also establishing a CAD cloud platform through collaborations with Japanese medical societies/informatics institutes~\cite{murao2020}. Leveraging the collected data, we are producing ready-to-use datasets for medical AI R\&D by 1) offering these datasets to AI firms, biopharma, and medical device makers and 2) using them as training/test data to develop customized AI solutions for such entities (Fig.~\ref{fig:business_ecosystem}). We aim to integrate Blockchain for data security and plan to synthesize rare disease data using generative AI.

\noindent \textbf{Contributions.} Our main contributions are as follows:
\vspace{-0.2cm}
\begin{itemize}
    \item \textbf{All-in-One Platform for Medical AI R\&D}: This is the first commercial medical imaging platform encompassing: 1) data collection, 2) data selection, 3) annotation, and 4) pre-processing.
    \vspace{-0.2cm}
    \item \textbf{Data Collected from Outside Europe and America}: Our platform collects under-represented\\CT/MRI/WSI data from Japan to mitigate the data imbalance.    
        \vspace{-0.2cm}
    \item \textbf{Blockchain-based Data Sharing}: We aim to share datasets $via$ Blockchain to avoid data leaks/tampering.
\end{itemize}

\section{METHODS} 
To address the issue of data imbalance in medical imaging, we established a system in which Japanese hospitals and clinics submit anonymized medical images, clinical data, and gene data, possibly with their annotation, to our cloud platform for commercial purposes. To expedite AI R\&D in this domain, we select, annotate (with an expert radiologist/pathologist's rigorous double-check), and pre-process these data to prepare ready-to-use datasets for medical AI R\&D.

\newpage

\subsection{Resolving Medical Imaging Data Imbalance}
\label{sec:medical_image_imbalance}

\begin{figure}[!t]
  \begin{center}
   \begin{tabular}{c}
   \includegraphics[width=16.5cm]{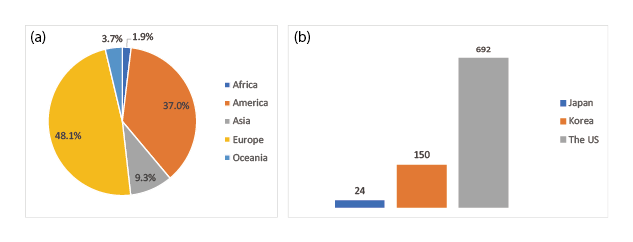}
   \end{tabular}
  \end{center}
   \caption[example] 
   { \label{fig:market_overview} 
(a) Percentage of imaging biobanks per continent~\cite{Gabelloni22}; (b) Count of AI/ML-based medical devices per country. Although recent data for Europe is unavailable, as of 2019, there were more CE-marked devices than FDA-cleared ones.}
   \vspace{5mm}
   \end{figure}

Medical imaging data imbalance is critical because non-representative data introduces bias and health inequity. Unfortunately, as shown in Fig.~\ref{fig:market_overview}, the majority of medical imaging data originates from Europe/America and less than 10\% comes from Asia, despite the Asian population comprising over 60\% of the global population. Moreover, most medical imaging databases are either commercially unavailable or available within the country.

Particularly In Japan, a stark lack of data available for commercial use has led to very few AI/ML-based medical devices (Fig.~\ref{fig:market_overview}). This is mainly attributed to 1) cultural apprehensions about data sharing, 2) diverse ethical codes across hospitals and clinics, and 3) varied data entry systems across institutions. To overcome these challenges and foster data sharing in Japan, and Asia more broadly, we have embarked on several initiatives:

\begin{itemize}
\item  Delivering invited talks at medical conferences and through mass media
\item Offering hospitals and clinics an anonymization tool equipped with data extraction features
\item Employing Blockchain techniques to circulate data, ensuring prevention against data leaks and tampering
\item Distributing the usage fee when the dataset is used for commercial purposes
\end{itemize}

   \begin{figure}[!t]
  \begin{center}
   \begin{tabular}{c}
   \includegraphics[width=16.5cm]{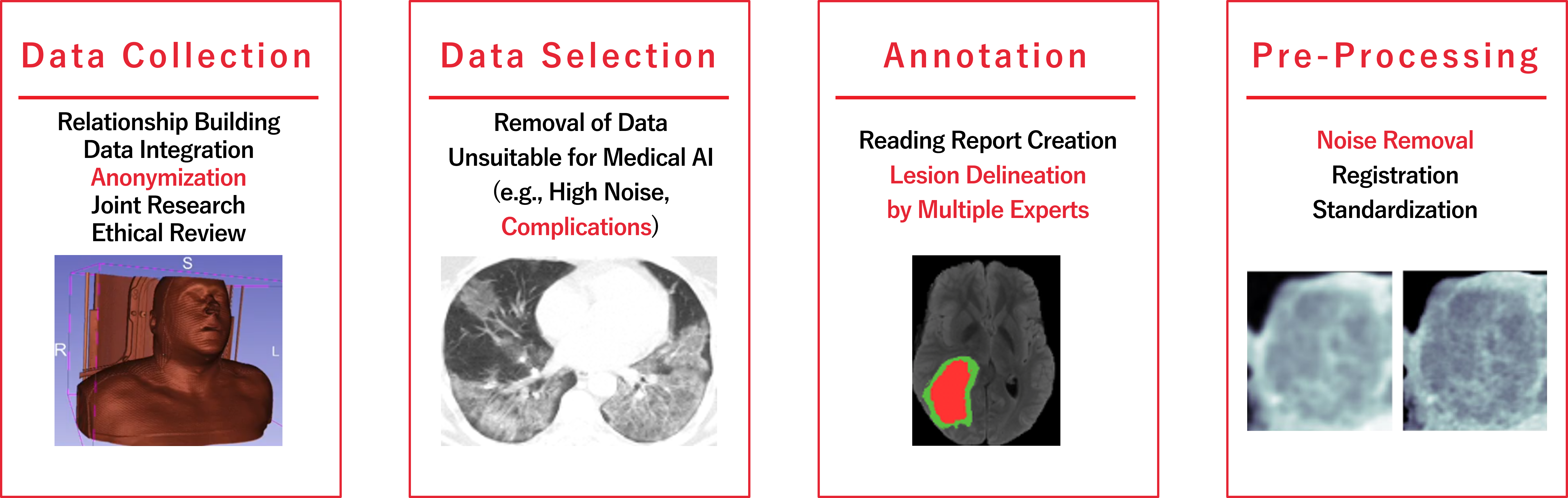}
   \end{tabular}
  \end{center}
   \caption[example] 
   { \label{fig:preparation_process} 
Preparation process of medical imaging datasets for AI training/test: 1) data collection~\cite{vallieres17}; 2) data selection~\cite{rampa2020}; 3) annotation~\cite{Cheng2019}; 4) pre-processing~\cite{deFarias22}.}
   \end{figure} 

\newpage

\subsection{All-in-One Platform for AI R\&D in Medical Imaging}
The preparation of medical imaging datasets for AI training/test involves four key steps: 1) data collection; 2) data selection; 3)
annotation; 4) pre-processing. Although several startups focus on specific steps---like data collection (e.g., Segmed and Gradient Health) or annotation (e.g., Centaur Labs)---no single platform exists that encompasses all four steps. This lack has led to the significant time and cost barriers globally for both medical AI R\&D and its translation to clinical practice. As Fig.~\ref{fig:preparation_process} shows, to fill this gap,  we aim to create an all-in-one platform covering CT, MRI, and WSI, along with clinical data and gene data, for accelerating medical AI and drug discovery AI innovations:

\begin{enumerate}
\item \textbf{Data Collection}: We foster relationships with hospitals/clinics, integrate data dispersed throughout the institution, anonymize imaging/clinical data, engage in joint AI research (which is typically mandatory in Japan), and undergo ethical review.
\item \textbf{Data Selection}: We exclude data unsuitable for medical AI, particularly data with excessive noise or complications.
\item \textbf{Annotation}: We delineate lesions/organ structures, ensuring a thorough double-check, and generate reading reports.
\item \textbf{Pre-processing}: We reduce image noise, align paired scans (e.g., MRI/CT pairs) or multiple acquisitions over time (e.g., time-series CT data), and standardize imaging/clinical/annotation data.

\end{enumerate}

\section{RESULTS} \label{sec:results}
We launched an all-in-one platform in Japan, the Asian nation with the world's highest aging population, to collect, process, and monetize medical images. Japan not only globally leads in the number of CT/MRI machines per capita but also boasts rich clinical data. We promoted our platform's concept to hospitals and clinics for dataset contributions and companies for dataset utilization in AI R\&D. To this end, we presented our vision at medical conferences, such as Japanese Society for Radiation Oncology and Japanese Society of Digital Pathology, and showcased it in the media, including a pitch on YouTube featuring top Japanese celebrities like Keisuke Honda and Yusuke Narita.

Many hospitals/clinics embraced our vision, generously offering their CT, MRI, and WSI cases (possibly with clinical data and gene data) for commercial use, totalling hundreds of thousands of cases. To further our mission Dr. Changhee Han, the founder of Callisto Inc., accepted roles as a specially-appointed associate professor at renowned university hospitals in Japan, including Osaka University's Medical Physics Lab and Nagasaki University's Pathology Informatics Lab, along with a director of Japanese Society of Digital Pathology.

Moreover, we are actively collaborating on medical AI and Blockchain research with major hospitals (e.g., Hiroshima University Hospital and Japanese Red Cross Suwa Hospital) and the University of Tokyo's Blockchain lab. These collaborations aim to propose clinically-beneficial AI models (e.g., generative AI for image prognosis) and integrate Blockchain (e.g., combining InterPlanetary Filesystem with Hyperledger Fabric) for secure data distribution while mitigating data breaches and tampering. We plan to 1) offer ready-to-use datasets to AI firms, biopharma, and medical device manufacturers, and 2) use them as training/test data to develop tailored medical AI/drug discovery AI solutions for such entities. We have already provided datasets upon requests, also engaging in customized medical AI/drug discovery development.

\section{CONCLUSION} 
To fast-track medical AI R\&D and its subsequent clinical implementation, we pioneered the first comprehensive platform encompassing: 1) data collection, 2) data selection, 3) annotation, and 4) pre-processing. Moreover, we aimed to address the data imbalance issue in medical imaging by commercially disseminating under-represented CT/MRI/WSI data from Japan, and Asia more broadly. Moving forward, we plan to synthesize hard-to-obtain data, such as rare disease ones, using generative AI---Dr. Changhee Han has published over 10 papers on generative medical imaging AI, covering data augmentation, super resolution, and unsupervised anomaly detection~\cite{han2021}.

\acknowledgments     

This work has been supported by Endowed Chair for Blockchain Innovation and the Mohammed bin Salman Center for Future Science and Technology for Saudi-Japan Vision 2030 (MbSC2030) at The University of Tokyo.


\bibliography{report}   
\bibliographystyle{spiebib}   

\end{document}